\documentclass[
]{ceurart}

\usepackage{listings}
\usepackage{subcaption}
\usepackage{url}
\begin{document}

\copyrightyear{2025}
\copyrightclause{Copyright for this paper by its authors.
  Use permitted under Creative Commons License Attribution 4.0
  International (CC BY 4.0).}

\conference{In: R. Campos, A. Jorge, A. Jatowt, S. Bhatia, M. Litvak (eds.): Proceedings of the Text2Story'25 Workshop, Lucca (Italy), 10-April-2025}

\title{Explainable AI Components for Narrative Map Extraction}


\author[1,2]{Brian Keith}[%
orcid=0000-0001-5734-8962,
email=brian.keith@ucn.cl,
url=https://briankeithn.github.io/,
]
\cormark[1]

\author[2]{Fausto German}[%
]

\author[3]{Eric Krokos}[%
]

\author[3]{Sarah Joseph}[%
]

\author[2]{Chris North}[%
]

\address[1]{Universidad Católica del Norte, Av. Angamos 0610, Antofagasta, 1270709, Chile}
\address[2]{Virginia Tech, 620 Drillfield Drive, Blacksburg, VA, 24061, USA}
\address[3]{U.S. Government, Washington, D.C. 20500, USA}

\cortext[1]{Corresponding author.}

\begin{abstract}
As narrative extraction systems grow in complexity, establishing user trust through interpretable and explainable outputs becomes increasingly critical. This paper presents an evaluation of an Explainable Artificial Intelligence (XAI) system for narrative map extraction that provides meaningful explanations across multiple levels of abstraction. Our system integrates explanations based on topical clusters for low-level document relationships, connection explanations for event relationships, and high-level structure explanations for overall narrative patterns. In particular, we evaluate the XAI system through a user study involving 10 participants that examined narratives from the 2021 Cuban protests. The analysis of results demonstrates that participants using the explanations made the users trust in the system's decisions, with connection explanations and important event detection proving particularly effective at building user confidence. Survey responses indicate that the multi-level explanation approach helped users develop appropriate trust in the system's narrative extraction capabilities. This work advances the state-of-the-art in explainable narrative extraction while providing practical insights for developing reliable narrative extraction systems that support effective human-AI collaboration.
\end{abstract}


\maketitle

\section{Introduction}
Understanding and extracting narratives from large collections of text documents presents significant challenges in natural language processing and visual analytics \cite{keith2023survey}. As narrative extraction methods become more sophisticated, particularly with the rise of complex Artificial Intelligence (AI) models, there is an increasing need to make these processes transparent and interpretable for users \cite{barredo2020xai,adadi2018xai}. This is especially crucial in domains such as journalism, intelligence analysis, and digital humanities, where analysts need not only to identify narratives but also to understand how and why specific narrative structures were extracted \cite{keith2023iui}.

Narrative maps---graph-based representations that capture the connections between events in a story---have emerged as a powerful tool for narrative sensemaking \cite{keith2020maps}. Throughout this work, we consider events---the basic unit of narratives---to be represented by single documents, following a document-based representation of news narratives \cite{keith2023survey}, under the assumption that a single news article usually contains a single main event \cite{norambuenaevaluating}.

Narrative maps are structures that represent events as nodes and their relationships as edges \cite{keith2020maps}, allowing analysts to explore how different parts of a narrative connect and evolve over time. However, the extraction of these maps often relies on complex pipelines involving multiple AI models \cite{keith2023iui}, from embedding generation to clustering and graph optimization. This complexity creates a ``black box'' effect \cite{wenskovitch2020interactive}, where users may not understand why certain events are connected or how the narrative structure was determined.

At the fundamental level, narrative extraction involves both low-level text processing and high-level structure generation, requiring explanations at different granularities \cite{keith2023iui}. The connections between events can be based on various factors \cite{keith2021vis} including temporal sequences, causal relationships, and thematic similarities, making it difficult to explain why specific relationships were identified. Furthermore, narrative extraction typically combines multiple AI models \cite{keith2023survey}, necessitating explanations that bridge different types of processing and help users understand how these components work together to produce the final narrative structure.

In this paper, we present an Explainable AI (XAI) system specifically designed for the task of narrative map extraction. Our approach provides explanations at multiple levels of the extraction process \cite{keith2023iui} through three main components. First, we leverage topical clusters \cite{kim2011topic,churchill2022topic} to generate keyword-based explanation for understanding the low-level document space. Second, we create a connection explanation framework that clarifies why events are linked in the narrative structure using explanations based on SHAP (Shapley Additive Explanations) values \cite{shap2017} and information about shared topics and entities between events. Third, we implement a high-level explanation system that provides descriptive names \cite{laban2017newslens} for storylines and identifies important events.

Our results from the user study demonstrate that providing explanations at multiple levels helps analysts develop appropriate \textit{trust}---understood in the broad sense of reliability, predictability, and efficiency \cite{hoffman2021measuring}---in extracted narrative structures, leading to more effective human-AI collaboration. This work addresses a critical gap in narrative extraction research by making these complex processes more transparent and interpretable to end users. Our system and data used to extract narratives is available in a public GitHub repository\footnote{\url{https://github.com/briankeithn/narrative-maps}}.

The remainder of this paper is organized as follows. Section 2 reviews related work. Section 3 presents our methodology, including the XAI components and the user study. Section 4 reports the results of the user study. Section 5 presents the discussion. Section 6 concludes with future directions.

\section{Related Work}
Research in explainable artificial intelligence for text analysis has emerged as a critical area as natural language processing systems become more complex. Adadi and Berrada \cite{adadi2018xai} provide a comprehensive survey of XAI approaches, categorizing them into model-agnostic and model-dependent techniques. Model-agnostic approaches can be applied to any machine learning model without consideration of internal structure, while model-dependent techniques are tailored to specific architectures. In text analysis, common XAI methods include simplification-based explanations \cite{grex}, relevance-based explanations \cite{shap2017}, and visual explanations \cite{LIME}.

The integration of XAI with visual analytics systems has received particular attention in recent research. Hohman et al. \cite{hohman2018visual} examine how visualization techniques can help reveal the inner workings of deep learning models for text processing. This work demonstrates that visual analytics can bridge the gap between complex AI models and human understanding. Building on this foundation, recent work by Vivacqua et al. \cite{vivacqua2020hci} specifically addresses XAI in the context of visualizations and sensemaking, showing how transparency in AI models can foster trust between humans and automated systems \cite{sevenHCIchallenges}.

In the domain of narrative understanding, XAI faces unique challenges due to the temporal and causal structure inherent in narratives. Abbott \cite{abbott2008cambridge} establishes that narratives have underlying temporal and causal structures that distinguish them from other forms of text. This structural complexity creates additional requirements for explanation systems. Keith and Mitra \cite{keith2020maps} introduce the concept of narrative maps as a representation for computational narrative extraction, demonstrating the need for explanations that can address both local event relationships and global narrative structure.

The explanation of narrative structures presents distinct challenges compared to general text analysis. Narrative extraction often relies on event-based models \cite{keith2023survey}, which must capture both the individual events and their interconnections. Traditional XAI approaches for text classification or similarity measurement must be adapted to account for these narrative-specific requirements. Recent work by Keith et al. \cite{keith2021vis} establishes design guidelines for narrative maps in sensemaking tasks, highlighting the importance of explanations that align with analysts' cognitive processes.

While existing research has made progress in explaining individual components of narrative analysis systems, there remains a gap in providing comprehensive explanations that span the entire extraction pipeline. Current approaches typically focus on either low-level text processing or high-level structure analysis, but rarely address both in an integrated manner. Our work builds upon these foundations while addressing the challenges of explaining narrative extraction across multiple levels of abstraction.

Finally, we note that, although explainable AI offers significant benefits for narrative extraction systems, we need to acknowledge potential limitations. XAI methods may introduce information overload when explanations are too technical or detailed, overwhelming users rather than helping to understand the AI model \cite{ferguson2022explanations}. There is also the risk that explanations function merely as proxies for complex underlying processes, potentially providing only a surface-level understanding that may not fully represent the actual computational mechanisms \cite{ribeiro2016why}. Additionally, misalignment between the outputs of the algorithm and the corresponding explanations can occur, leading to decreased trust if the behavior of the system contradicts its explanations \cite{ZERILLI2022100455}. In our approach, we address these concerns by designing explanations at appropriate levels of abstraction, focusing on providing general pointers for understanding the big picture while also supporting specific confirmatory tasks, such as verifying connection validity between events.

\section{Methodology}
\subsection{Problem Definition and Assumptions}
As narrative extraction systems grow in complexity, users face increasing difficulty understanding both how these systems work and why specific narrative structures were extracted. This paper addresses the specific problem of providing meaningful explanations across multiple levels of abstraction in narrative map extraction systems to enhance transparency, user trust, and effective human-AI collaboration. 

In this context, our approach is built on several key assumptions. First, users must understand both the low-level relationship between documents and the high-level narrative structures to develop the appropriate trust in the system. Second, different types of explanations are required for different aspects of the system, such as topical clusters, connections, and storylines. Third, explanations bridging the gap between computational processes and human cognition should enhance trust and usability. Additionally, explanations should balance detail with comprehensibility, avoiding information overload while providing sufficient insight to support user understanding and decision making. 

These assumptions guided our development of our multi-level explanation components that address different aspects of the narrative extraction process while maintaining cognitive accessibility.

\subsection{Overview of XAI System for Narrative Maps}
Our explainable AI system for narrative maps addresses the challenge of providing meaningful explanations across multiple levels of abstraction in the narrative extraction process. We note that our XAI system builds upon previous extraction models and interactive prototypes \cite{keith2020maps,keith2023iui} and thus we do not explain the underlying components or the extraction method of the narrative maps in detail. In particular, the system integrates with a mixed multi-model pipeline \cite{keith2023iui} that combines low-level continuous spaces for document representation with high-level discrete structures for narrative visualization.

In general, the extraction process takes news articles as input and proceeds in two main phases: extraction and post-processing \cite{keith2023iui}. During extraction, the system maps articles into an embedding space, computes coherence between events using information about topical clusters and similarity measures, and uses linear programming to build the optimal narrative structure \cite{keith2020maps}. Post-processing then simplifies this structure, following design guidelines \cite{keith2022design} to make it more understandable by removing redundant connections while preserving the core narrative elements.

At the foundation of our implementation lies the principle that explanations must bridge the gap between the computational processes of narrative extraction and the cognitive processes of human analysts. We accomplish this through a three-tiered approach that provides explanations for the document space (a low-level model without structure), the narrative structure (a high-level model that captures the underlying narrative connections), and the \textit{connection} between these two. These explanations are generated through a combination of model-agnostic and model-dependent techniques, carefully selected to maintain computational efficiency while providing meaningful insights. 

\begin{figure}[!htb]
\centering
\includegraphics[width=\textwidth]{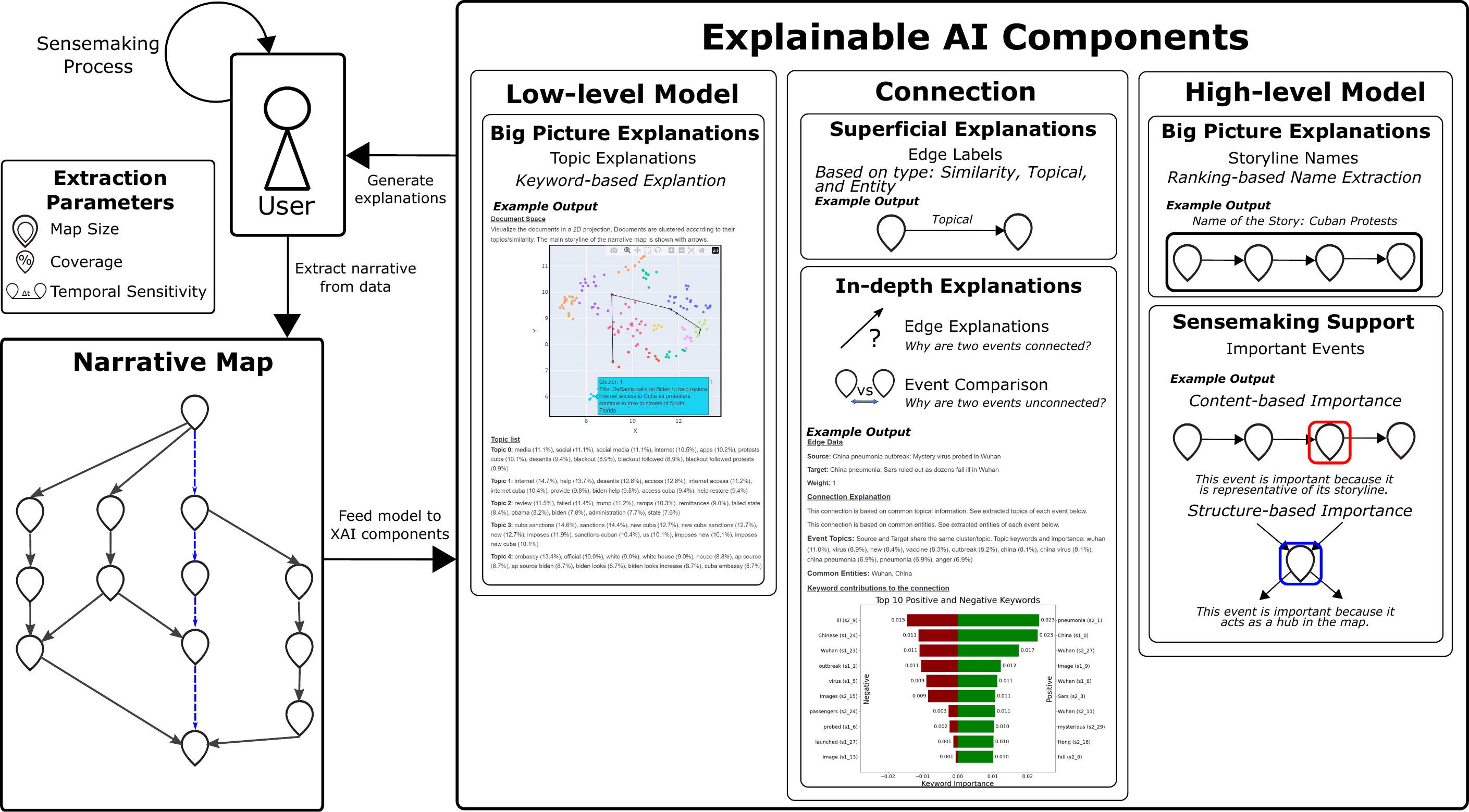}
\caption{Implemented pipeline showing the extraction of narrative map and explainable AI components to aid users in understanding the underlying model, including topic clusters, connection labels, and storyline names.}
\label{fig:pipeline}
\end{figure}

We show the pipeline of our system in Figure \ref{fig:pipeline}. To use the system, the user extracts a narrative map representation from data with user-defined parameters (map size, story coverage, and temporal sensitivity). The narrative representation is then fed to the XAI components. The XAI components focus on each of the previously mentioned tiers (the low-level model, the high-level model, and the connection between these two).

For the low-level model, the XAI system uses keyword representations of topical clusters to provide big picture explanations that seek to capture general patterns in the document space. For the high-level model, the XAI system uses storyline names to provide big picture explanations. The storyline names are extracted from relevant parts of the documents using a ranking-based name extraction algorithm. The XAI system also provides supporting explanations for sensemaking purposes. In particular, the system identifies important events to help users identify highly relevant documents. These important events are identified based on the relevance of their content with respect to their storyline or whether they are relevant to the overall structure of the narrative (e.g., acting as a central node in the graph).

Finally, for the connection between the two models, the XAI system assigns labels to each connection, which provide a super explanation of the type of connection. The types identified by the XAI system are based on a previous taxonomy used by analysts during the narrative sensemaking process \cite{keith2022design}. While these labels provide a general view of why two specific events were connected, they do not provide sufficient details. Thus, we expand upon each relevant element of the different types of connections, including a breakdown of the \textit{topic} to which each event belongs, the common \textit{entities} they have, and the specific contributions of each \textit{keyword} towards similarity. The information shown depends on the specific label assigned to the connection (e.g., a connection with an ``Entity'' label will display the common entities). For the topical information, we specify whether the events share common topics and the keywords that define these topics. Following the same format from the explanations of the low-level space. For common entities, we simply intersect the sets of identified entities and display them accordingly. To find the keywords with the highest contribution (in positive or negative terms), we use the SHAP library \cite{shap2017} with a permutation-based approach.

\subsection{Low-level Space Explanations}
The low-level space explanations seek to help analysts understand the document embedding space and the topical relationships between events. We implement this by identifying topical clusters to reveal the underlying structure of the document space. In particular, we use HDBSCAN clustering \cite{mcinnes2017hdbscan} to identify coherent groups of documents in the embedding space. This hierarchical clustering method was used because it has proven to work reliably for text data in the context of narrative extraction \cite{keith2020maps,keith2023iui}. Furthermore, it can be used to extract cluster probability vectors (i.e. soft clustering), allowing for a more nuanced assignment of events to each cluster \cite{mcinnes2017hdbscan}. 

For each cluster, we generate keyword-based explanations using a modified TF-IDF (Term Frequency and Inverse Document Frequency) representation \cite{debole2004supervised} that incorporates both global and local term importance, allowing us to capture cluster-specific terminology while maintaining context from the broader document collection. Specifically, we compute the importance score $S$ for term $t$ in cluster $c$ as:

\begin{equation}
S(t,c) = TF(t,c) \cdot IDF_{global}(t) \cdot IDF_{local}(t,c)
\end{equation}

where $TF(t,c)$ represents the term frequency in the cluster, $IDF_{global}(t)$ captures the term's importance across the entire corpus, and $IDF_{local}(t,c)$ measures the term's specificity within the cluster.

The cluster visualization provides a spatial view of the document relationships through a 2D projection using UMAP \cite{mcinnes2018umap}. This projection preserves both local and global structure, allowing analysts to see how documents relate to each other within and across topic clusters. This visualization includes interactive tooltips that display the cluster membership and key terms for specific documents.

\begin{figure}[!htb]
    \centering
    \includegraphics[width=0.48\textwidth]{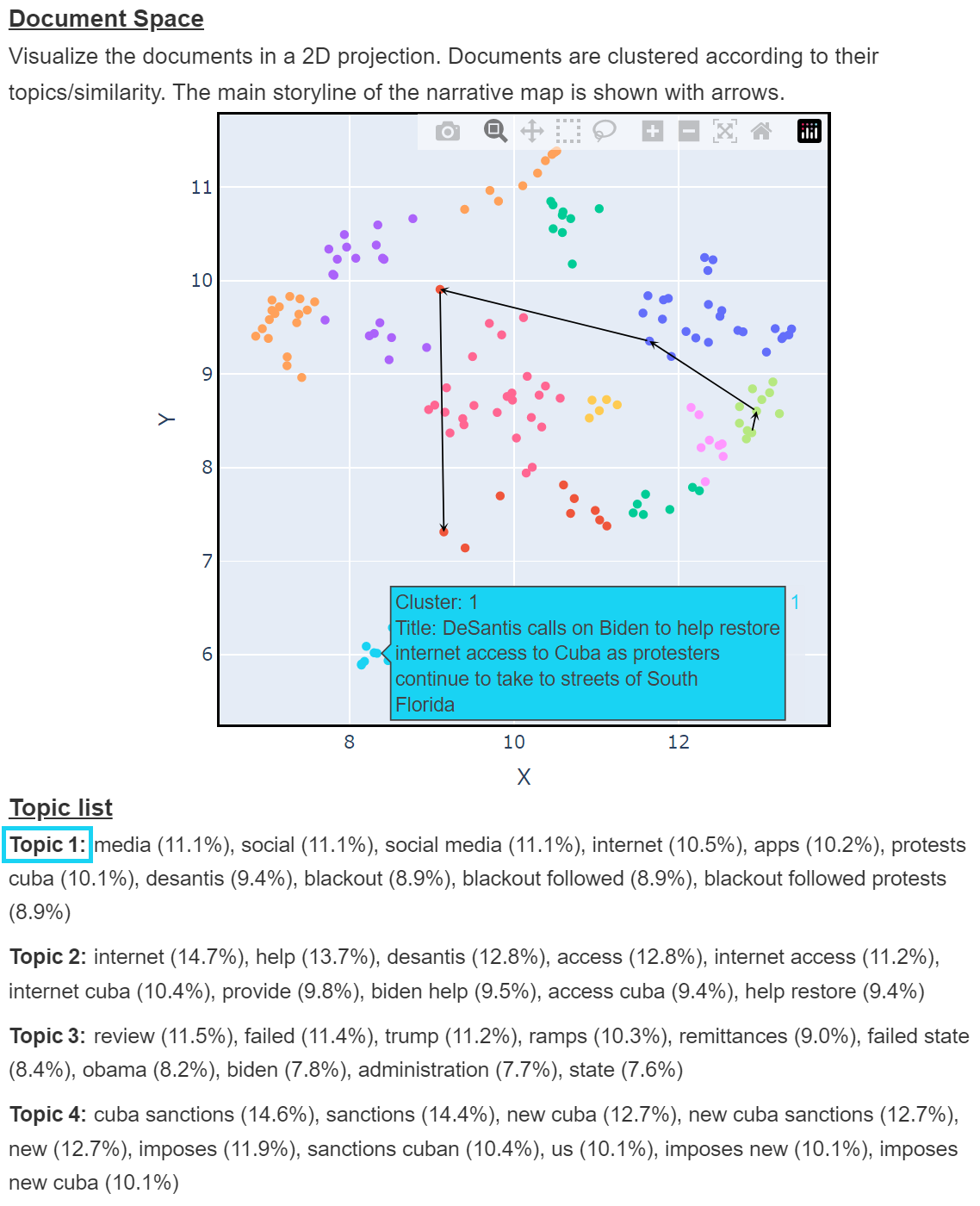}
    \hfill
    \includegraphics[width=0.48\textwidth]{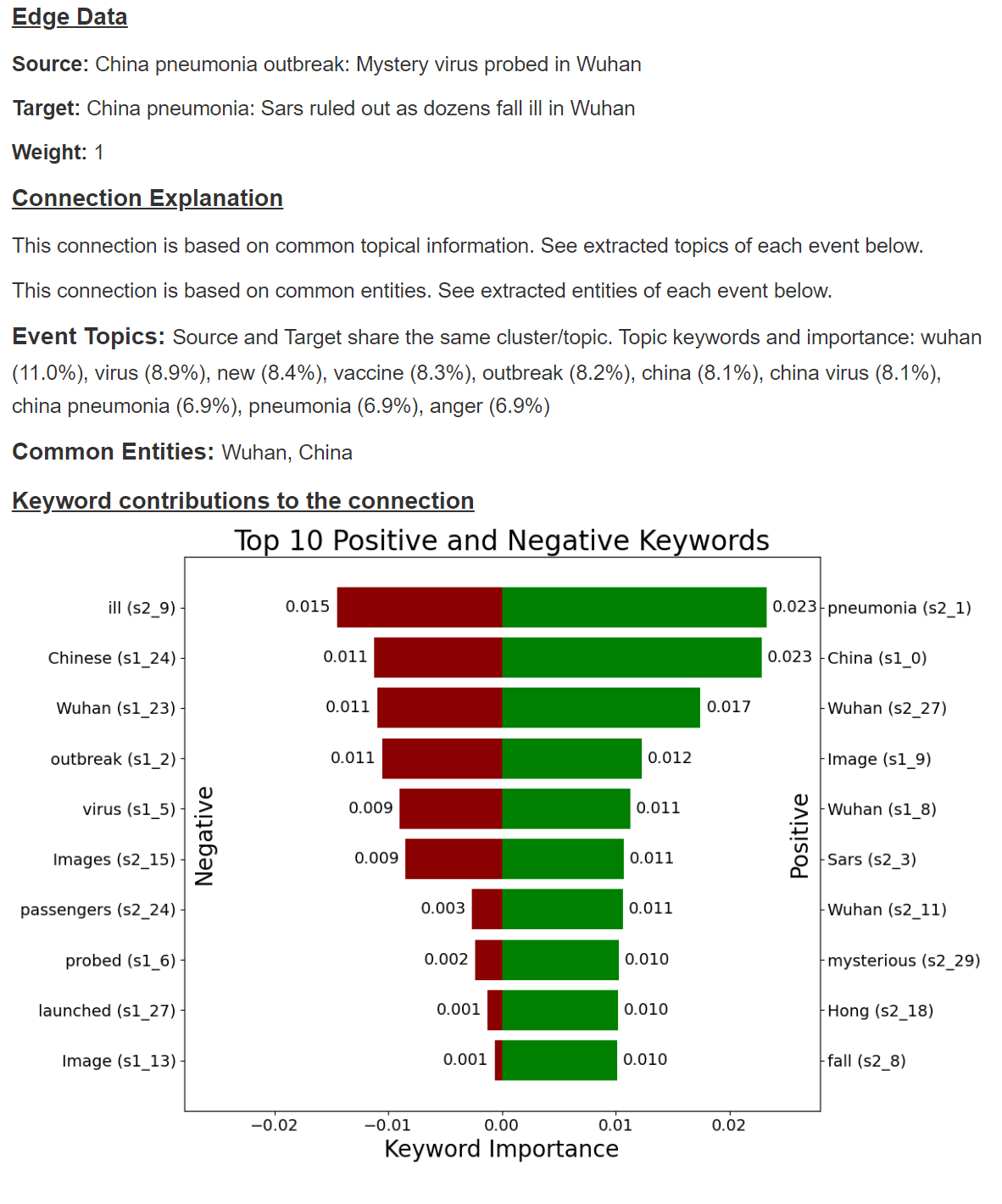}
    
    \begin{minipage}[t]{0.48\textwidth}
        \centering
        \textbf{(a)}
    \end{minipage}
    \hfill
    \begin{minipage}[t]{0.48\textwidth}
        \centering
        \textbf{(b)}
    \end{minipage}
    
    \caption{Example explanations using a COVID-19 dataset (not used in the tasks of the user study, shown only as an illustration). \textbf{(a)} Topical cluster explanations for the low-level space, including a scatter plot of the space and an overlay of the main storyline. Tooltips with the same color as the corresponding cluster are also included. In this figure, we also highlighted the corresponding description on the topic list. \textbf{(b)} Explanation for an edge between two events. This explanation specifies the type of connection (e.g., topical or entity-based), the topics of the events, and the keyword contributions that directly impact their similarity and thus the coherence score.}
    \label{fig:xai_examples}
\end{figure}

To ensure the explanations remain interpretable, we limit the number of keywords shown for each cluster based on empirical testing with analysts. The system displays the top-k keywords where k is determined dynamically based on the cluster size and keyword importance distribution. This approach prevents information overload while ensuring that sufficient context is provided for understanding each topic cluster. We show the topic explanations in Figure \ref{fig:xai_examples}(a).

\subsection{Connection Explanations}
The connection explanation component addresses the critical task of explaining why specific events are connected in the narrative. Our approach generates these explanations through three interconnected processes: connection label generation, detailed explanation generation, and event comparison.

Connection label generation relies on a taxonomy derived from narrative map design guidelines \cite{keith2022design}. The system classifies connections into three primary types: similarity-based, entity-based, and topical connections. To determine the connection type, we analyze the components of the coherence measure used in the extraction process. The system computes the relative contribution of clustering similarity versus basic text similarity, assigning the label ``Topical'' when the clustering component contributes more than 50\% to the coherence score, and ``Similarity'' otherwise. Entity-based connections are identified through named entity recognition, with an additional overlap score based on Jaccard similarity to handle partial entity matches:

\begin{equation}
    overlap(e_1, e_2) = \frac{|tokens(e_1) \cap tokens(e_2)|}{|tokens(e_1) \cup tokens(e_2)|}
\end{equation}

For detailed explanation generation, we implement a model-agnostic approach using SHAP values \cite{shap2017} to identify the most influential terms contributing to event connections. The system generates explanations by analyzing both positive and negative contributions to the connection strength. We modify the standard SHAP implementation to produce interpretable explanations by focusing on the headline and first thirty words of each event, which typically contain the most relevant information in news narratives \cite{norambuenaevaluating}. We show the connection explanation function in Figure \ref{fig:xai_examples}(b).

The event comparison functionality extends these explanation capabilities to help analysts understand why certain events are not connected. This component applies the same analysis techniques used for connected events but focuses on explaining the factors that resulted in events remaining unconnected.

\subsection{High-level Structure Explanations}
High-level structure explanations focus on making the global narrative structure interpretable through automated storyline naming and important event detection. These explanations help analysts understand the broader narrative patterns while maintaining connection to the underlying evidence.

\subsubsection{Storyline Name Extraction}
The storyline naming process builds upon the work of Laban and Hearst \cite{laban2017newslens} in timeline summarization. Our approach identifies candidate names by extracting maximal noun phrases that contain both proper nouns and abstract terms. We show a simplfiied, but illustrative version of the model to score these candidates using a linear combination of factors:

\begin{equation}
    \label{eq:3}
    Score(name) = \alpha \cdot C_{entity} + \beta \cdot C_{abstract} + \gamma \cdot C_{coverage} - \delta \cdot O_{overlap}
\end{equation}

where $C_{entity}$, $C_{abstract}$, and $C_{coverage}$ represent the presence of entities, abstract terms, and coverage of storyline content respectively, while $O_{overlap}$ penalizes redundancy with existing storyline names. The weights $\alpha$, $\beta$, $\gamma$, and $\delta$ are determined empirically.

\subsubsection{Important Event Detection}
Important event detection combines both content-based and structural approaches to identify key events in the narrative. For content importance, we compute the similarity between each event's embedding and the centroid of its storyline. Structural importance is assessed through degree centrality in the narrative graph, weighted by the coherence values of the connections. The system identifies important events by selecting the top-n events according to each criterion:

\begin{equation}
    I_{content}(e) = cos(e, centroid_{storyline})
\end{equation}
\begin{equation}
    I_{structure}(e) = \sum_{v \in N(e)} coherence(e,v)
\end{equation}

The explanation system visually marks these important events on the narrative map. Events that score highly on both content and structural measures receive additional emphasis, as they represent key narrative elements that are both thematically central and well-connected.

\subsection{Evaluation of the System}
We evaluated our system through a user study focused on analyzing narratives about the 2021 Cuban protests. Using an insight-based evaluation methodology \cite{north2011comparison}, we assessed how effectively our XAI components supported narrative sensemaking tasks. The dataset used in this user study comprised 160 news articles from diverse sources, providing comprehensive coverage while maintaining manageable computational requirements. We recruited 10 participants with backgrounds in computer science, communications, and national security. The participants reported minimal prior knowledge of the 2021 Cuban protests, with a mean (M) \textit{familiarity} of 1.4 on a 5-point Likert scale and a standard deviation (SD) of 0.52. The participants first received a 15-minute training session using a separate COVID-19 news dataset \cite{keith2022design} to familiarize themselves with the features of the system, ensuring that they could focus on evaluating the XAI features rather than learning the mechanics of the system.

The insight-based evaluation required users to obtain as many insights as possible with our system. The insights were then categorized and counted. This open-ended task gives analysts freedom to explore the dataset and provides an approximation of a realistic narrative sensemaking task. Finally, we asked participants to complete a follow-up questionnaire (see Table \ref{tab:likert}) on their perception of the XAI components to help develop an understanding of the narrative structure and verify their analysis.

\begin{table}[t]
\centering
\caption{Post-task questionnaire with Likert scale questions.}
\label{tab:likert}
\resizebox{\linewidth}{!}{%
\small
\begin{tabular}{p{2.75cm}p{14cm}}
\toprule
\textbf{Category} & \textbf{Questionnaire Item} \\
\midrule
{\textbf{General}} & \textbf{Usefulness:} The explanations provided by the system were useful. \\
 & \textbf{Trust:} The explanations made me trust the system more. \\
\midrule
{\textbf{Storyline Names}} & \textbf{Correctness:} The storyline names provided by the system were appropriate for each storyline. \\
 & \textbf{Relevance:} The storyline names provided by the system were relevant. \\
 & \textbf{Usefulness:} The storyline names provided by the system made the narrative easier to understand. \\
\midrule
{\textbf{Connections}} & \textbf{Labels Correctness:} The connection labels provided by the system about the types of connections made sense. \\
 & \textbf{Labels Usefulness:} The connection labels provided by the system about the types of connections were useful to understand the connections. \\
 & \textbf{Connection Explanations:} The in-depth explanations for connections provided by the system were useful. \\
 & \textbf{Event Comparison:} The event comparison explanations for unconnected events provided by the system were useful. \\
\midrule
{\textbf{Important Events}} & \textbf{Relevance:} The important events selected by the system were relevant. \\
 & \textbf{Usefulness:} The important events selected by the system made the narrative easier to understand. \\
\bottomrule
\end{tabular}%
}
\end{table}

\section{Evaluation Results}
\begin{figure}[!htb]
    \centering
    \includegraphics[width=\columnwidth]{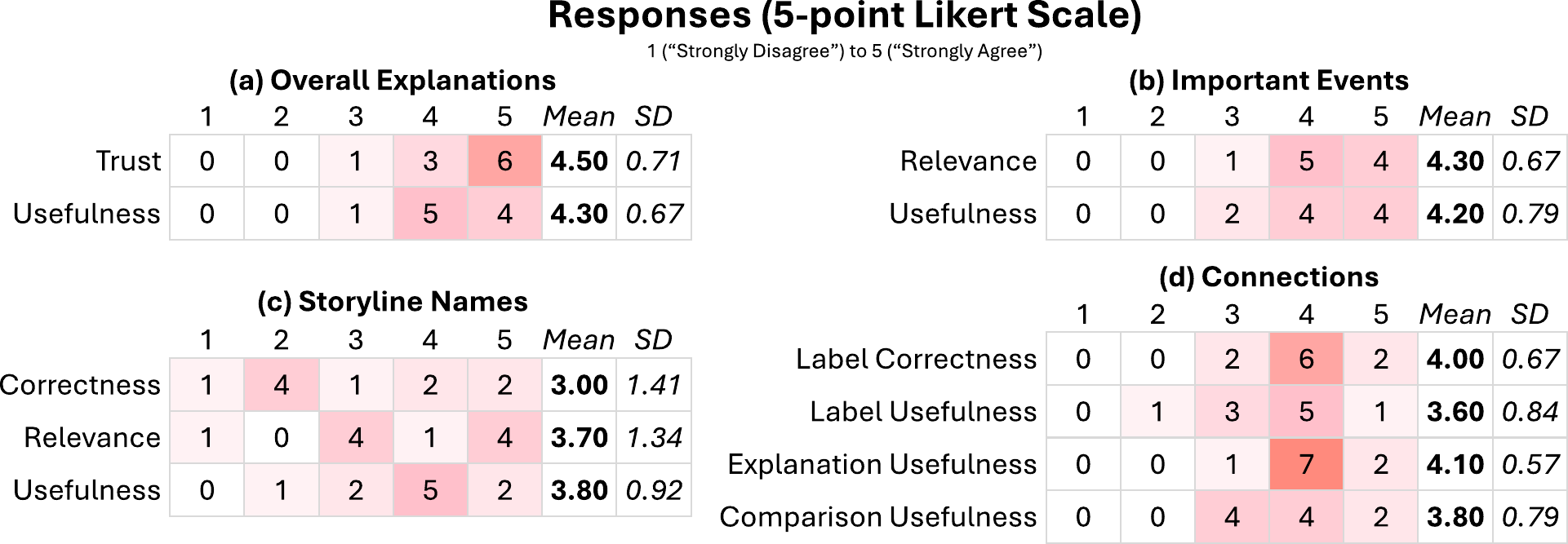}
    \caption{A tally of the answers to our survey questions on the \textit{explainable AI} components.}
    \label{fig:tally_i3}
\end{figure}

Our evaluation focused on user perception and trust in the system's XAI components. We show the results in Figure \ref{fig:tally_i3}. The results are reported on a 5-point Likert scale.

\textbf{Overall Explanations.} Survey results indicated that, in general, the explanation components significantly increased user trust in the system (M = 4.5, SD = 0.71) and were considered useful by the participants (M = 4.3, SD = 0.67). In particular, the participants reported that the explanations of the topical clusters using TF-IDF increased confidence in understanding the clustering of documents.

\textbf{Important Events.} The important events detected by the system were considered relevant (M = 4.3, SD = 0.67) and useful (M = 4.2, SD = 0.79). Compared to other components of the system (story names and connection labels), important event detection proved to be more consistently valuable, helping participants quickly identify key narrative elements.

\textbf{Storyline Names.} These high-level structure explanations showed mixed results in building trust. Automated storyline naming received more variable feedback in terms of correctness, presenting a higher standard deviation and lower mean (M = 3.0, SD = 1.41), and only slightly better in terms of relevance (M = 3.7, SD = 1.34) and usefulness (M = 3.8, SD = 0.92). Paraphrased from a participant: ``\textit{Storyline names are not always correct, but they are relevant and useful.}'' 

\textbf{Connections.} Label-based explanations demonstrated varying levels of trust enhancement among participants. The generated labels were generally considered correct (M = 4.0, SD = 0.67), but slightly less useful (M = 3.6, SD = 0.84). The SHAP-based keyword explanations helped the participants verify connection validity, increasing confidence in the system's linking decisions, and were considered generally useful (M = 4.1, SD = 0.57). The comparison tool was considered slightly less useful (M = 3.8, SD = 0.79). Analysis of participant feedback revealed that the combination of explanations at multiple levels enhanced trust in the system's narrative extraction capabilities.

These results suggest that our XAI system effectively supports user trust in narrative analysis systems, particularly when explanations span multiple levels of abstraction. The variable effectiveness of different explanation types indicates opportunities for future refinement of explanation strategies to better align with user expectations and trust-building needs.

\section{Discussion and Limitations}
\subsection{Explainable AI Results}
Our results indicate that XAI components substantially support narrative sensemaking tasks when properly integrated into the analysis workflow. In particular, the participants of our user study considered the addition of XAI beneficial and it increased their trust in the underlying models, even if the explanations themselves were not always particularly useful for the assigned task. However, our user evaluation also shows that there is some contention on the usefulness of each individual component, such as storyline names, which were sometimes considered very useful and sometimes discarded as mostly useless, depending on the participant.

For most use cases, XAI provides useful scaffolding and enables users to potentially find more insights. However, developing appropriate XAI methods that capture the intricacies of the underlying models and are useful from a user's perspective is a complex task. In general, based on our results, we recommend designing such explanations with the goal of helping users understand the big picture by providing general pointers and providing support for specific confirmatory tasks, such as checking whether an element of the model is valid (e.g., event connections). In our evaluation, methods that provided too much information or were too technical in nature were generally regarded negatively, as participants were likely overwhelmed by the information overload of the explanations \cite{ferguson2022explanations}. 

Our findings provide some practical insights for future narrative extraction systems. In particular, the variable effectiveness of different explanation types demonstrates that explanation strategies should be tailored to specific components of the narrative structure, rather than applying a single approach uniformly. Furthermore, our analysis of results suggests that a design approach that combines multiple explanation modalities while prioritizing cognitive accessibility over technical comprehensiveness. In general, we found that explanations contributed to user-perceived trust, which underscores the importance of transparency in AI-assisted analytical tasks.

\subsection{Trust Building with Explainable AI}
The proposed explainable AI framework relies on a set of machine learning and data mining algorithms to interpret the semantic representations embedded in the narrative maps. While this method provides a surface-level explanation of the outcomes of the model, it introduces a significant challenge to the reliability and validity of the generated outcomes. Minor adjustments at each stage of the process can yield disparate outputs, thereby raising concerns about the consistency and accuracy of the results. However, from the results of our user study, despite these potential limitations, the users still found the system to be useful for solving sensemaking tasks.

Similarly, XAI methods often function as proxies that attempt to clarify the impact or contribution of various features within the model to increase transparency \cite{ribeiro2016why}. For example, in the case of our narrative map explanations, we leverage topical clusters and storyline name extraction to summarize the events and provide a general overview. However, even if these explanations may only capture a surface-level understanding of the processes governing the system, our user study shows that these explanations are still useful, even if they do not fully represent the underlying model.

Moreover, a key concern when implementing XAI methods is the potential risk of providing inaccurate or misleading explanations, which, in turn, could lead to a decrease in user trust in the system \cite{Manzey2012, Dzindolet2003, Dietvorst2020} and the underlying AI model \cite{ZERILLI2022100455}. This is an issue that should be carefully addressed to ensure that XAI methods are transparent, trustworthy, and reliable. Thus, to ensure that users are aware that the algorithm outputs may not always be perfectly aligned with the XAI components, future versions of the system can incorporate warning mechanisms or indicators that can signal when the explanations provided by the model might be less reliable or accurate.

When it comes to the storyline names, text summarization is an active area of research \cite{WIDYASSARI20221029} with many challenges and opportunities, particularly in handling context, maintaining coherence, and ensuring accuracy while condensing large volumes of information. While the current method follows an extractive strategy, a shift towards an \textit{abstractive} strategy might be more effective \cite{Lin_Ng_2019}. This approach would involve creating names with words that might not directly exist in the storylines but accurately describe them. Implementing this could be achieved with generative neural networks \cite{Gao2020}, which can distill the essential aspects of the storyline.

\subsection{Limitations}
First, we note that we did not compare the proposed system with a proper baseline in our user study, such as another system from the literature or the same system without XAI. Instead, our focus was on exploring how participants used the features and whether they considered them useful. The findings of this study could help inform future versions of such XAI systems to provide better user support. 

Furthermore, we note that participants had no easy way to determine whether the extracted narratives were indeed correct, as the system does not provide evaluation metrics on the factual accuracy of the narratives. However, evaluating the correctness of narratives in general is an open problem, as there are no unified metrics that work in all cases \cite{keith2020maps}.

Scalability remains a consideration for larger datasets. Although our implementation handled the test dataset effectively, the computational complexity of generating comprehensive explanations may become prohibitive for larger narrative collections. Future work could explore hierarchical explanation strategies that can scale more effectively to larger datasets while maintaining explanation quality.

Despite these limitations, our results demonstrate the value of integrated XAI components in narrative analysis systems. The positive user feedback on explanation utility suggests that our approach effectively supports the sensemaking process of analysts. Future work could address these limitations while expanding XAI methods to handle additional types of narrative relationships and explanation needs.

\section{Conclusions}
This paper presented the evaluation of an XAI system to explain AI-driven narrative extraction systems through multiple levels of abstraction. Our approach bridges the gap between low-level text processing and high-level narrative structures, providing analysts with meaningful explanations that enhance trust in automated narrative analysis. Through empirical evaluation, we have shown that integrated explainable AI components improve user confidence in working with complex narrative structures.

The results of our user study indicate that narrative maps augmented with XAI techniques effectively build user trust in automated analysis systems. The combination of topical cluster explanations, connection explanations, and high-level structure explanations provides analysts with a coherent understanding of system decisions at multiple levels. Our evaluation demonstrates that connection type labels and important event detection enhance user confidence, while more complex explanations require additional refinement to maximize their trust-building potential.

Several promising directions emerge for future research. The development of more sophisticated temporal and causal explanation strategies could further enhance user trust in automated narrative extraction. In addition, the exploration of adaptive explanation approaches that respond to different levels of user expertise and trust requirements presents an important avenue for investigation.

The broader implications of this work extend beyond narrative analysis to the general challenge of building trustworthy AI systems. Our findings suggest that carefully designed explanations at multiple levels of abstraction can effectively support human-AI collaboration by establishing appropriate levels of trust. As narrative analysis systems continue to evolve, the principles and approaches developed in this work can inform the design of future explainable AI systems that users can confidently rely upon for complex analytical tasks.

\begin{acknowledgments}
This research was supported by industry, government, and institute members of the NSF SHREC Center, which was founded in the IUCRC program of the National Science Foundation. Brian Keith is also supported by Project 202311010033-VRIDT-UCN.
\end{acknowledgments}

\section*{Declaration on Generative AI}
During the preparation of this work, the authors used Grammarly and Writefull integrated with Overleaf to perform grammar and spelling corrections. After using these tools/services, the authors reviewed and edited the content as needed and take full responsibility for the publication’s content. 

\bibliography{sample-ceur}

\begin{thebibliography}{33}
\expandafter\ifx\csname natexlab\endcsname\relax\def\natexlab#1{#1}\fi
\providecommand{\url}[1]{\texttt{#1}}
\providecommand{\href}[2]{#2}
\providecommand{\path}[1]{#1}
\providecommand{\DOIprefix}{doi:}
\providecommand{\ArXivprefix}{arXiv:}
\providecommand{\URLprefix}{URL: }
\providecommand{\Pubmedprefix}{pmid:}
\providecommand{\doi}[1]{\href{http://dx.doi.org/#1}{\path{#1}}}
\providecommand{\Pubmed}[1]{\href{pmid:#1}{\path{#1}}}
\providecommand{\bibinfo}[2]{#2}
\ifx\xfnm\relax \def\xfnm[#1]{\unskip,\space#1}\fi
\bibitem[{Keith~Norambuena et~al.(2023)Keith~Norambuena, Mitra, and North}]{keith2023survey}
\bibinfo{author}{B.~F. Keith~Norambuena}, \bibinfo{author}{T.~Mitra}, \bibinfo{author}{C.~North},
\newblock \bibinfo{title}{A survey on event-based news narrative extraction},
\newblock \bibinfo{journal}{ACM Comput. Surv.} \bibinfo{volume}{55} (\bibinfo{year}{2023}). \URLprefix \url{https://doi.org/10.1145/3584741}. \DOIprefix\doi{10.1145/3584741}.
\bibitem[{{Barredo Arrieta} et~al.(2020){Barredo Arrieta}, Díaz-Rodríguez, {Del Ser}, Bennetot, Tabik, Barbado, Garcia, Gil-Lopez, Molina, Benjamins, Chatila, and Herrera}]{barredo2020xai}
\bibinfo{author}{A.~{Barredo Arrieta}}, \bibinfo{author}{N.~Díaz-Rodríguez}, \bibinfo{author}{J.~{Del Ser}}, \bibinfo{author}{A.~Bennetot}, \bibinfo{author}{S.~Tabik}, \bibinfo{author}{A.~Barbado}, \bibinfo{author}{S.~Garcia}, \bibinfo{author}{S.~Gil-Lopez}, \bibinfo{author}{D.~Molina}, \bibinfo{author}{R.~Benjamins}, \bibinfo{author}{R.~Chatila}, \bibinfo{author}{F.~Herrera},
\newblock \bibinfo{title}{Explainable artificial intelligence (xai): Concepts, taxonomies, opportunities and challenges toward responsible ai},
\newblock \bibinfo{journal}{Information Fusion} \bibinfo{volume}{58} (\bibinfo{year}{2020}) \bibinfo{pages}{82--115}.
\bibitem[{Adadi and Berrada(2018)}]{adadi2018xai}
\bibinfo{author}{A.~Adadi}, \bibinfo{author}{M.~Berrada},
\newblock \bibinfo{title}{Peeking inside the black-box: A survey on explainable artificial intelligence (xai)},
\newblock \bibinfo{journal}{IEEE Access} \bibinfo{volume}{6} (\bibinfo{year}{2018}) \bibinfo{pages}{52138--52160}.
\bibitem[{Keith~Norambuena et~al.(2023)Keith~Norambuena, Mitra, and North}]{keith2023iui}
\bibinfo{author}{B.~F. Keith~Norambuena}, \bibinfo{author}{T.~Mitra}, \bibinfo{author}{C.~North},
\newblock \bibinfo{title}{Mixed multi-model semantic interaction for graph-based narrative visualizations},
\newblock in: \bibinfo{booktitle}{Proceedings of the 28th International Conference on Intelligent User Interfaces}, IUI '23, \bibinfo{publisher}{Association for Computing Machinery}, \bibinfo{address}{New York, NY, USA}, \bibinfo{year}{2023}, p. \bibinfo{pages}{866–888}. \URLprefix \url{https://doi.org/10.1145/3581641.3584076}. \DOIprefix\doi{10.1145/3581641.3584076}.
\bibitem[{Keith~Norambuena and Mitra(2020)}]{keith2020maps}
\bibinfo{author}{B.~Keith~Norambuena}, \bibinfo{author}{T.~Mitra},
\newblock \bibinfo{title}{Narrative maps: An algorithmic approach to represent and extract information narratives},
\newblock in: \bibinfo{booktitle}{Proc. ACM Hum.-Comput. Interact.}, volume~\bibinfo{volume}{4}, \bibinfo{publisher}{ACM}, \bibinfo{address}{New York, NY, USA}, \bibinfo{year}{2020}, p. \bibinfo{pages}{33 pages}.
\bibitem[{Keith~Norambuena et~al.(2020)Keith~Norambuena, Horning, and Mitra}]{norambuenaevaluating}
\bibinfo{author}{B.~Keith~Norambuena}, \bibinfo{author}{M.~Horning}, \bibinfo{author}{T.~Mitra},
\newblock \bibinfo{title}{Evaluating the inverted pyramid structure through automatic 5w1h extraction and summarization},
\newblock in: \bibinfo{booktitle}{Proc. of the 2020 Computation + Journalism Symposium}, \bibinfo{publisher}{C + J 2020}, \bibinfo{address}{Boston, MA, USA}, \bibinfo{year}{2020}, pp. \bibinfo{pages}{1--7}.
\bibitem[{Wenskovitch and North(2020)}]{wenskovitch2020interactive}
\bibinfo{author}{J.~Wenskovitch}, \bibinfo{author}{C.~North},
\newblock \bibinfo{title}{Interactive artificial intelligence: Designing for the" two black boxes" problem},
\newblock \bibinfo{journal}{Computer} \bibinfo{volume}{53} (\bibinfo{year}{2020}) \bibinfo{pages}{29--39}.
\bibitem[{Keith~Norambuena et~al.(2021)Keith~Norambuena, Mitra, and North}]{keith2021vis}
\bibinfo{author}{B.~F. Keith~Norambuena}, \bibinfo{author}{T.~Mitra}, \bibinfo{author}{C.~North},
\newblock \bibinfo{title}{Narrative sensemaking: Strategies for narrative maps construction},
\newblock in: \bibinfo{booktitle}{2021 IEEE Visualization Conference (VIS)}, \bibinfo{publisher}{IEEE}, \bibinfo{address}{New Orleans, LA, USA}, \bibinfo{year}{2021}, pp. \bibinfo{pages}{181--185}.
\bibitem[{Kim and Oh(2011)}]{kim2011topic}
\bibinfo{author}{D.~Kim}, \bibinfo{author}{A.~Oh},
\newblock \bibinfo{title}{Topic chains for understanding a news corpus},
\newblock in: \bibinfo{editor}{A.~Gelbukh} (Ed.), \bibinfo{booktitle}{Computational Linguistics and Intelligent Text Processing}, \bibinfo{publisher}{Springer Berlin Heidelberg}, \bibinfo{address}{Berlin, Heidelberg}, \bibinfo{year}{2011}, pp. \bibinfo{pages}{163--176}.
\bibitem[{Churchill and Singh(2022)}]{churchill2022topic}
\bibinfo{author}{R.~Churchill}, \bibinfo{author}{L.~Singh},
\newblock \bibinfo{title}{The evolution of topic modeling},
\newblock \bibinfo{journal}{ACM Comput. Surv.} \bibinfo{volume}{54} (\bibinfo{year}{2022}).
\bibitem[{Lundberg and Lee(2017)}]{shap2017}
\bibinfo{author}{S.~M. Lundberg}, \bibinfo{author}{S.-I. Lee},
\newblock \bibinfo{title}{A unified approach to interpreting model predictions},
\newblock in: \bibinfo{editor}{I.~Guyon}, \bibinfo{editor}{U.~V. Luxburg}, \bibinfo{editor}{S.~Bengio}, \bibinfo{editor}{H.~Wallach}, \bibinfo{editor}{R.~Fergus}, \bibinfo{editor}{S.~Vishwanathan}, \bibinfo{editor}{R.~Garnett} (Eds.), \bibinfo{booktitle}{Advances in Neural Information Processing Systems}, volume~\bibinfo{volume}{30}, \bibinfo{publisher}{Curran Associates, Inc.}, \bibinfo{address}{Long Beach, CA, USA}, \bibinfo{year}{2017}. \URLprefix \url{https://proceedings.neurips.cc/paper_files/paper/2017/file/8a20a8621978632d76c43dfd28b67767-Paper.pdf}.
\bibitem[{Laban and Hearst(2017)}]{laban2017newslens}
\bibinfo{author}{P.~Laban}, \bibinfo{author}{M.~A. Hearst},
\newblock \bibinfo{title}{newslens: building and visualizing long-ranging news stories},
\newblock in: \bibinfo{booktitle}{Proc. of the Events and Stories in the News Workshop}, \bibinfo{publisher}{ACL}, \bibinfo{address}{Vancouver, Canada}, \bibinfo{year}{2017}, pp. \bibinfo{pages}{1--9}.
\bibitem[{Hoffman et~al.(2021)Hoffman, Mueller, Klein, and Litman}]{hoffman2021measuring}
\bibinfo{author}{R.~Hoffman}, \bibinfo{author}{S.~Mueller}, \bibinfo{author}{G.~Klein}, \bibinfo{author}{J.~Litman},
\newblock \bibinfo{title}{Measuring trust in the xai context}  (\bibinfo{year}{2021}).
\bibitem[{Konig et~al.(2008)Konig, Johansson, and Niklasson}]{grex}
\bibinfo{author}{R.~Konig}, \bibinfo{author}{U.~Johansson}, \bibinfo{author}{L.~Niklasson},
\newblock \bibinfo{title}{G-rex: A versatile framework for evolutionary data mining},
\newblock in: \bibinfo{booktitle}{2008 IEEE International Conference on Data Mining Workshops}, \bibinfo{publisher}{IEEE}, \bibinfo{address}{Pisa, Italy}, \bibinfo{year}{2008}, pp. \bibinfo{pages}{971--974}.
\bibitem[{Ribeiro et~al.(2016)Ribeiro, Singh, and Guestrin}]{LIME}
\bibinfo{author}{M.~T. Ribeiro}, \bibinfo{author}{S.~Singh}, \bibinfo{author}{C.~Guestrin},
\newblock \bibinfo{title}{"why should i trust you?": Explaining the predictions of any classifier},
\newblock in: \bibinfo{booktitle}{Proc. of the 22nd ACM SIGKDD International Conference on Knowledge Discovery and Data Mining}, KDD '16, \bibinfo{publisher}{ACM}, \bibinfo{address}{New York, NY, USA}, \bibinfo{year}{2016}, p. \bibinfo{pages}{1135–1144}.
\bibitem[{Hohman et~al.(2018)Hohman, Kahng, Pienta, and Chau}]{hohman2018visual}
\bibinfo{author}{F.~Hohman}, \bibinfo{author}{M.~Kahng}, \bibinfo{author}{R.~Pienta}, \bibinfo{author}{D.~H. Chau},
\newblock \bibinfo{title}{Visual analytics in deep learning: An interrogative survey for the next frontiers},
\newblock \bibinfo{journal}{IEEE trans. on visualization and computer graphics} \bibinfo{volume}{25} (\bibinfo{year}{2018}) \bibinfo{pages}{2674--2693}.
\bibitem[{Vivacqua et~al.(2020)Vivacqua, Stelling, Garcia, and Gouvea}]{vivacqua2020hci}
\bibinfo{author}{A.~S. Vivacqua}, \bibinfo{author}{R.~Stelling}, \bibinfo{author}{A.~C.~B. Garcia}, \bibinfo{author}{L.~C. Gouvea},
\newblock \bibinfo{title}{Explanations and sensemaking with ai and hci},
\newblock in: \bibinfo{booktitle}{Proc. of the IX Latin American Conference on Human Computer Interaction}, CLIHC '19, \bibinfo{publisher}{ACM}, \bibinfo{address}{New York, NY, USA}, \bibinfo{year}{2020}.
\bibitem[{Stephanidis et~al.(2019)Stephanidis, Salvendy, of~the Group Margherita~Antona, Chen, Dong, Duffy, Fang, Fidopiastis, Fragomeni, Fu, Guo, Harris, Ioannou, ah~(Kate)~Jeong, Konomi, Krömker, Kurosu, Lewis, Marcus, Meiselwitz, Moallem, Mori, Nah, Ntoa, Rau, Schmorrow, Siau, Streitz, Wang, Yamamoto, Zaphiris, and Zhou}]{sevenHCIchallenges}
\bibinfo{author}{C.~C. Stephanidis}, \bibinfo{author}{G.~Salvendy}, \bibinfo{author}{M.~of~the Group Margherita~Antona}, \bibinfo{author}{J.~Y.~C. Chen}, \bibinfo{author}{J.~Dong}, \bibinfo{author}{V.~G. Duffy}, \bibinfo{author}{X.~Fang}, \bibinfo{author}{C.~Fidopiastis}, \bibinfo{author}{G.~Fragomeni}, \bibinfo{author}{L.~P. Fu}, \bibinfo{author}{Y.~Guo}, \bibinfo{author}{D.~Harris}, \bibinfo{author}{A.~Ioannou}, \bibinfo{author}{K.~ah~(Kate)~Jeong}, \bibinfo{author}{S.~Konomi}, \bibinfo{author}{H.~Krömker}, \bibinfo{author}{M.~Kurosu}, \bibinfo{author}{J.~R. Lewis}, \bibinfo{author}{A.~Marcus}, \bibinfo{author}{G.~Meiselwitz}, \bibinfo{author}{A.~Moallem}, \bibinfo{author}{H.~Mori}, \bibinfo{author}{F.~F.-H. Nah}, \bibinfo{author}{S.~Ntoa}, \bibinfo{author}{P.-L.~P. Rau}, \bibinfo{author}{D.~Schmorrow}, \bibinfo{author}{K.~Siau}, \bibinfo{author}{N.~Streitz}, \bibinfo{author}{W.~Wang}, \bibinfo{author}{S.~Yamamoto}, \bibinfo{author}{P.~Zaphiris}, \bibinfo{author}{J.~Zhou},
\newblock \bibinfo{title}{Seven hci grand challenges},
\newblock \bibinfo{journal}{International Journal of Human–Computer Interaction} \bibinfo{volume}{35} (\bibinfo{year}{2019}) \bibinfo{pages}{1229--1269}.
\bibitem[{Abbott(2008)}]{abbott2008cambridge}
\bibinfo{author}{H.~P. Abbott}, \bibinfo{title}{The Cambridge introduction to narrative}, \bibinfo{publisher}{Cambridge University Press}, \bibinfo{address}{One Liberty Plaza, New York, NY, USA}, \bibinfo{year}{2008}.
\bibitem[{Ferguson et~al.(2022)Ferguson, Franklin, and Lagnado}]{ferguson2022explanations}
\bibinfo{author}{A.~N. Ferguson}, \bibinfo{author}{M.~Franklin}, \bibinfo{author}{D.~Lagnado},
\newblock \bibinfo{title}{Explanations that backfire: Explainable artificial intelligence can cause information overload},
\newblock in: \bibinfo{booktitle}{Proc. of the Annual Meeting of the Cognitive Science Society}, volume~\bibinfo{volume}{44}, \bibinfo{publisher}{Cognitive Science Society}, \bibinfo{address}{Sydney, NSW, Australia}, \bibinfo{year}{2022}.
\bibitem[{Ribeiro et~al.(2016)Ribeiro, Singh, and Guestrin}]{ribeiro2016why}
\bibinfo{author}{M.~T. Ribeiro}, \bibinfo{author}{S.~Singh}, \bibinfo{author}{C.~Guestrin}, \bibinfo{title}{"why should i trust you?": Explaining the predictions of any classifier}, \bibinfo{year}{2016}. \href{http://arxiv.org/abs/1602.04938}{{\tt arXiv:1602.04938}}.
\bibitem[{Zerilli et~al.(2022)Zerilli, Bhatt, and Weller}]{ZERILLI2022100455}
\bibinfo{author}{J.~Zerilli}, \bibinfo{author}{U.~Bhatt}, \bibinfo{author}{A.~Weller},
\newblock \bibinfo{title}{How transparency modulates trust in artificial intelligence},
\newblock \bibinfo{journal}{Patterns} \bibinfo{volume}{3} (\bibinfo{year}{2022}) \bibinfo{pages}{100455}. \URLprefix \url{https://www.sciencedirect.com/science/article/pii/S2666389922000289}. \DOIprefix\doi{https://doi.org/10.1016/j.patter.2022.100455}.
\bibitem[{Keith~Norambuena et~al.(0)Keith~Norambuena, Mitra, and North}]{keith2022design}
\bibinfo{author}{B.~F. Keith~Norambuena}, \bibinfo{author}{T.~Mitra}, \bibinfo{author}{C.~North},
\newblock \bibinfo{title}{Design guidelines for narrative maps in sensemaking tasks},
\newblock \bibinfo{journal}{Information Visualization} \bibinfo{volume}{21} (\bibinfo{year}{0}) \bibinfo{pages}{220--245}.
\bibitem[{McInnes et~al.(2017)McInnes, Healy, and Astels}]{mcinnes2017hdbscan}
\bibinfo{author}{L.~McInnes}, \bibinfo{author}{J.~Healy}, \bibinfo{author}{S.~Astels},
\newblock \bibinfo{title}{hdbscan: Hierarchical density based clustering},
\newblock \bibinfo{journal}{Journal of Open Source Software} \bibinfo{volume}{2} (\bibinfo{year}{2017}) \bibinfo{pages}{205}.
\bibitem[{Debole and Sebastiani(2003)}]{debole2004supervised}
\bibinfo{author}{F.~Debole}, \bibinfo{author}{F.~Sebastiani},
\newblock \bibinfo{title}{Supervised term weighting for automated text categorization},
\newblock in: \bibinfo{booktitle}{Proceedings of the 2003 ACM Symposium on Applied Computing}, SAC '03, \bibinfo{publisher}{Association for Computing Machinery}, \bibinfo{address}{New York, NY, USA}, \bibinfo{year}{2003}, p. \bibinfo{pages}{784–788}. \URLprefix \url{https://doi.org/10.1145/952532.952688}. \DOIprefix\doi{10.1145/952532.952688}.
\bibitem[{McInnes et~al.(2018)McInnes, Healy, and Melville}]{mcinnes2018umap}
\bibinfo{author}{L.~McInnes}, \bibinfo{author}{J.~Healy}, \bibinfo{author}{J.~Melville}, \bibinfo{title}{Umap: Uniform manifold approximation and projection for dimension reduction}, \bibinfo{howpublished}{arXiv preprint arXiv:1802.03426}, \bibinfo{year}{2018}.
\bibitem[{North et~al.(2011)North, Saraiya, and Duca}]{north2011comparison}
\bibinfo{author}{C.~North}, \bibinfo{author}{P.~Saraiya}, \bibinfo{author}{K.~Duca},
\newblock \bibinfo{title}{A comparison of benchmark task and insight evaluation methods for information visualization},
\newblock \bibinfo{journal}{Information Visualization} \bibinfo{volume}{10} (\bibinfo{year}{2011}) \bibinfo{pages}{162--181}.
\bibitem[{Manzey et~al.(2012)Manzey, Reichenbach, and Onnasch}]{Manzey2012}
\bibinfo{author}{D.~Manzey}, \bibinfo{author}{J.~Reichenbach}, \bibinfo{author}{L.~Onnasch},
\newblock \bibinfo{title}{Human performance consequences of automated decision aids},
\newblock \bibinfo{journal}{Journal of Cognitive Engineering and Decision Making} \bibinfo{volume}{6} (\bibinfo{year}{2012}) \bibinfo{pages}{57--87}. \DOIprefix\doi{10.1177/1555343411433844}.
\bibitem[{Dzindolet et~al.(2003)Dzindolet, Peterson, Pomranky, Pierce, and Beck}]{Dzindolet2003}
\bibinfo{author}{M.~T. Dzindolet}, \bibinfo{author}{S.~A. Peterson}, \bibinfo{author}{R.~A. Pomranky}, \bibinfo{author}{L.~G. Pierce}, \bibinfo{author}{H.~P. Beck},
\newblock \bibinfo{title}{The role of trust in automation reliance},
\newblock \bibinfo{journal}{International Journal of Human-Computer Studies} \bibinfo{volume}{58} (\bibinfo{year}{2003}) \bibinfo{pages}{697--718}. \DOIprefix\doi{10.1016/S1071-5819(03)00038-7}.
\bibitem[{Dietvorst and Bharti(2020)}]{Dietvorst2020}
\bibinfo{author}{B.~J. Dietvorst}, \bibinfo{author}{S.~Bharti},
\newblock \bibinfo{title}{People reject algorithms in uncertain decision domains because they have diminishing sensitivity to forecasting error},
\newblock \bibinfo{journal}{Psychological Science} \bibinfo{volume}{31} (\bibinfo{year}{2020}) \bibinfo{pages}{1302--1314}. \DOIprefix\doi{10.1177/0956797620948841}.
\bibitem[{Widyassari et~al.(2022)Widyassari, Rustad, Shidik, Noersasongko, Syukur, Affandy, and Setiadi}]{WIDYASSARI20221029}
\bibinfo{author}{A.~P. Widyassari}, \bibinfo{author}{S.~Rustad}, \bibinfo{author}{G.~F. Shidik}, \bibinfo{author}{E.~Noersasongko}, \bibinfo{author}{A.~Syukur}, \bibinfo{author}{A.~Affandy}, \bibinfo{author}{D.~R. I.~M. Setiadi},
\newblock \bibinfo{title}{Review of automatic text summarization techniques \& methods},
\newblock \bibinfo{journal}{Journal of King Saud University - Computer and Information Sciences} \bibinfo{volume}{34} (\bibinfo{year}{2022}) \bibinfo{pages}{1029--1046}. \URLprefix \url{https://www.sciencedirect.com/science/article/pii/S1319157820303712}. \DOIprefix\doi{https://doi.org/10.1016/j.jksuci.2020.05.006}.
\bibitem[{Lin and Ng(2019)}]{Lin_Ng_2019}
\bibinfo{author}{H.~Lin}, \bibinfo{author}{V.~Ng},
\newblock \bibinfo{title}{Abstractive summarization: A survey of the state of the art},
\newblock \bibinfo{journal}{Proceedings of the AAAI Conference on Artificial Intelligence} \bibinfo{volume}{33} (\bibinfo{year}{2019}) \bibinfo{pages}{9815--9822}. \URLprefix \url{https://ojs.aaai.org/index.php/AAAI/article/view/5056}. \DOIprefix\doi{10.1609/aaai.v33i01.33019815}.
\bibitem[{Gao et~al.(2020)Gao, Wang, Liu, Guo, and Huang}]{Gao2020}
\bibinfo{author}{Y.~Gao}, \bibinfo{author}{Y.~Wang}, \bibinfo{author}{L.~Liu}, \bibinfo{author}{Y.~Guo}, \bibinfo{author}{H.~Huang},
\newblock \bibinfo{title}{Neural abstractive summarization fusing by global generative topics},
\newblock \bibinfo{journal}{Neural Computing and Applications} \bibinfo{volume}{32} (\bibinfo{year}{2020}) \bibinfo{pages}{5049--5058}. \URLprefix \url{https://doi.org/10.1007/s00521-018-3946-7}. \DOIprefix\doi{10.1007/s00521-018-3946-7}.

\end{thebibliography}


\end{document}